\documentclass[review]{elsarticle}

\usepackage{lineno,hyperref}
\modulolinenumbers[10]
\journal{Journal of \LaTeX\ Templates}
\usepackage[left=2.5cm, right=2.5cm, top=2cm]{geometry}
\usepackage{graphicx}
\usepackage{amsmath}
\usepackage{wrapfig}
\usepackage{float}
\usepackage{array}
\usepackage{booktabs}
\usepackage{color}
\usepackage{multirow}
\usepackage{caption}
\usepackage{rotating}
\usepackage{blindtext,pdflscape}
\begin{document}

\begin{frontmatter}

\title{I-vector Based Features Embedding for Heart Sound Classification}
\author[mymainaddress]{Mohammad Adiban}
\author[mysecondaryaddress]{Bagher BabaAli}
\author[mytertiaryaddress]{Saeedreza Shehnepoor}
\address[mymainaddress]{Sharif University of Technology, Iran}
\address[mysecondaryaddress]{University of Tehran, Iran}
\address[mytertiaryaddress]{University of Western Australia, Australia}

\begin{abstract}
Cardiovascular Disease (CVD) is considered as one of the principal causes of death in the world. 
Over recent years, this field of study has attracted researchers' attention to investigate heart sounds' patterns for disease diagnostics. In this study, an approach is proposed for normal/abnormal heart sound classification on the Physionet challenge 2016 dataset.
For the first time, a fixed length feature vector; called i-vector; is extracted from each heart sound 
using Mel Frequency Cepstral Coefficient (MFCC) features. Afterwards, Principal Component Analysis (PCA) transform and Variational Autoencoder (VAE) are applied on the i-vector to achieve dimension reduction.
Eventually, the reduced size vector is fed to Gaussian Mixture Models (GMMs) and Support Vector Machine (SVM) for classification purpose. Experimental results demonstrate the proposed method could achieve a performance improvement of 16\% based on Modified Accuracy (MAcc) compared with the baseline system on the Physionet2016 dataset.
\end{abstract}
\end{frontmatter}

\section{Introduction}
Cardiovascular disease (CVD) is one of the most common causes of death around the world and the leading cause of disability. Based on the information provided by the World Heart Association, 2017, 17.7 million people die every year due to CVD, equal to 31\% of all global deaths. The most prevalent CVDs are heart attacks and strokes. In 2013, all 194 members of the World Health Organization agreed to implement the Global Action Plan for the Prevention and Control of Non-communicable Diseases, a plan for 2013 to 2020, to be prepared against CVDs. Implementation of nine global and voluntary goals in this plan led to a significant drop in the number of premature deaths due to non-communicable diseases. \par
Accordingly, in recent years, researchers have showed a considerable interest in detecting heart diseases based on heart sounds ~\cite{ref2}. Most approaches in this context rely on sound segmentation and feature extraction. Extracted features are then fed to machine learning methods to simulate the system performance on real-world datasets. In addition, various studies are conducted for normal/abnormal heart sound classification using segmentation methods. Methods in this field can be categorized into three groups; Segmentation-based approaches, Wavelet-based approaches, and Time-Frequency based approaches. Methods in the first category, focus on using variable window of short time Fourier transform (S-transform), Hilbert transform, etc. in order to segment each audio and then apply different classification to detect hear normal/abnormal behavior \cite{ref4}\cite{ref5}\cite{ref6}\cite{ref7}. The second category is based on the wavelet transformation. Wavelet features are then fed to the well-known approaches such as SVM \cite{ref14}\cite{ref15}. Finally, recent approaches are concered with using Time-Frequency features such as Discrete Fourier Transform (DFT), MFCC, and so on. To determine each test sample class, varied classification approaches such as Convolutional Neural Network (CNN), Artificial Neural Network (ANN), etc., were adopted. \cite{ref13}\cite{ref17}\cite{ref18}\cite{ref19}. \par
Although approaches in this field have made remarkable progress, still they are not able to yield the desired performance. One reason for this issue is the paucity of systems that extract appropriate speech features from heart sounds. Hence, representing appropriate features is the most essential step toward improving the performance, although it is not achievable without using a suitable classifier. In the proposed framework, i-vector is used as a feature representation technique. The underlying motivation for using i-vector in this context is that human heart sounds can be considered as the physiological traits of a person~\cite{ref32} and only irregular events such as accidents, illnesses, genetic defects, or aging can alter or destroy these traits~\cite{ref32}. As a result, heart sounds are prone to being introduced by representative features like i-vectors. In the classification stage, GMMs and SVM are employed which are persuaded for modeling i-vector with mixture nature. In addition, feature reduction techniques such as PCA and VAE are also applied blueto achieve a compact and informative representation of the extracted i-vectors which lead to having discriminative features. So our contributions are summarized as follow:  \par
\begin{itemize}
    \item While i-vector is employed predominantly in various speech-related tasks, it is less known to the biosignal processing fields. In this work, the i-vector that is generally used for speaker recognition is adopted to the heart sound classification task.
    \item PCA and VAE are employed to reduce the size of i-vector and also extract the most significant features that lead to a more discriminative representation for next stage which is classification.
    \item Finally, the proposed approach outperforms the previous studies in terms of Specificity (Sp), Sensitivity (Se) and accuracy. Physionet 2016 pertinent studies are listed in \cite{Clifford_2017} and the best reported results belong to Potes \textit{et al.} \cite{Potes2016EnsembleOF}. Our proposed method outperforms their results in terms of Sp and Macc. \end{itemize}
In the following sections, first the related works are presented, then the proposed framework are described. Next, the experimental setup is discussed and then the results are analyzed in the experimental results section. Finally, the conclusion section concludes the whole research study. 
\section{Related Studies}
{A review of current approaches is presented in \cite{Dwivedi2019AlgorithmsFA}. Each presented approach is systematically reviewed and existing approaches are analyzed based on their performance. In the following section, a brief discussion about approaches in this field will be presented, which categorizes them into two subcategories.} 
\subsection{Segmentation Based Approaches}
Approaches in this category use features and audio segments to classify normal/abnormal heart sounds. 
In a study by~\cite{ref4}, an approach was proposed for automatic segmentation, using Hilbert transform. Features for this study included envelops near the peaks of S1, S2, the transmission points T12 from S1 to S2, and vice versa. Database for this study consisted of 7730s of heart sound from pathological patients, 600s from normal subjects, and finally 1496.8 s from Michigan MHSDB database. The average accuracy for sound with mixed S1, and S2 was 96.69\%, and it was reported 97.37\% for those with separated S1 and S2.\par
CNN based segmentation is proposed in \cite{renna2018convolutional}, which is common in image processing tasks, to segment heart sounds into their main components. The same concept is used in \cite{noman2019short} which incorporates CNN to extract short segment features from 1-dimensional, e.g. raw heart sound signal, and 2-dimensional, e.g. time-frequency representation of heart sound to segment these signals.
Another envelope extraction method was employed for heart sound segmentation is called Cardiac Sound Characteristic Waveform (CSCW). The work presented in~\cite{ref5} used this method for only a small set of heart sounds, including 9 sound recordings and 99.0\% accuracy was reported. No train-test split was performed for evaluation in this study. The work in~\cite{ref6} achieved an accuracy of 92.4\% for S1 and 93.5\% for S2 segmentation by engaging homomorphic filtering and Hidden Markov Model (HMM), on the PASCAL database~\cite{ref7}.
\subsection{Wavelet Based Approaches}
Wavelet-based approaches employ the wavelet transform to extract features, and then these extracted features are subsequently used for classification. In~\cite{ref3} the Shannon energy envelops for the local spectrum are calculated by a new method, which uses S-transform for every sound produced by the heart sound signals. Sensitivity and positive predictivity were evaluated on 80 heart sound recordings (including 40 normal and 40 pathological), and their values were reported over 95\%. The work investigated in~\cite{ref8} also adopted the same approach with wavelet analysis on the same database and accuracy was reported 90.9\% for S1 segmentation and this value was 93.3\% for S2 segmentation. The work in~\cite{ref14} also conducted a study to classify normal and pathological cases using Least Square Support Vector Machine (LSSVM) engaging wavelet to extract features. They evaluated their method on a dataset with heart sound of 64 patients (32 cases for train and 32 cases for test set) and reported 86.72\% for accuracy. In a work~\cite{ref15} with the same classifier, wavelet packets and extracted features are engaged like sample entropy and energy fraction as input. The dataset used for this problem consisted of 40 normal individuals and 67 pathological patients and they resulted in 97.17\% accuracy, 93.48\% sensitivity and 98.55\% specificity. Another study~\cite{ref16}, also used LSSVM as classifier while using the tunable-Q wavelet transform as input features. Evaluation in this study showed 98.8\% sensitivity and 99.3\% specificity on a dataset comprising 4628 cycles from 163 heart sound recordings, with the unknown number of patients. 
Fractional Fourier transform is proposed for feature extraction in \cite{abduh2019classification} and the extracted features are subsequently classified by a stacked autoencoder which yields a performance accuracy of 95\%.

A study on the expected duration of heart sound using HMM and Hidden Semi-Markov Model (HSMM) was introduced in~\cite{ref9}. In this study, positions of S1 and S2 sounds were initially labeled in 113 recordings. Afterwards, they calculated Gaussian distributions for the expected duration of each four states including S1, systole, S2, and diastole, using the average duration of mentioned sound and also autocorrelation analysis of systolic and diastolic duration. Homomorphic envelope plus three other frequency features (in 25-50, 50-100 and 100-150 Hz ranges) were among features they used for this study. Then they calculated Gaussian distributions for training HMM states and emission probabilities. Finally, for the decoding process, the backward and forward Viterbi algorithm was engaged. Sensitivity and Specificity were reported 98.8\% and 98.6\%, respectively. This work also proposed HSMM alongside logistic regression (for emission probability estimation) to accurately segment noisy, and real-world heart sound recording~\cite{ref10}. This work also used Viterbi algorithm to decode state sequences. For evaluation, they used a database of 10172s of heart sounds recorded from 112 patients. F1 score for this study is reported to be 95.63\%, improving over the previous state of the art study with 86.28\% on the same test set.\par
Other studies were also developed using other methods based on the feature extraction and classification using machine learning classifiers such as ANN, SVM, HMM, and K-Nearest Neighbor (KNN). For the distinction between spectral energy between normal and pathological recordings, the work introduced in~\cite{ref11} extracted five frequency bands and their spectral energy was given as input to ANN. Results on a dataset with 50 recorded sounds reveal 95\% sensitivity and 93.33\% specificity.
In a study by~\cite{ref12}, a discrete wavelet transform as well as a fuzzy logic was used for a three-class problem; including normal, pulmonary stenosis, and mitral stenosis. An ANN was employed to classify a dataset of 120 subjects with 50/50 split for train and test set. Reported results were 100\% for sensitivity, 95.24\% for specificity, and 98.33\% for average accuracy. Moreover, they used time-frequency as input for ANN in~\cite{ref13}. This work reported 90.4\% sensitivity, 97.44\% specificity, and 95\% accuracy on the same dataset for the same problem (three-class classification including normal, pulmonary and mitral stenosis heart valve diseases).\par

HMM was used by~\cite{ref18} to fit on the frequency spectrum from the heart cycle and used four HMMs for evaluating the posterior probability of the features given to model for classification. For better results, they used PCA as reduction procedure and results reported 95\% sensitivity, 98.8\% specificity, and 97.5\% accuracy on a dataset with 60 samples. 
The KNN was used in ~\cite{ref19} on the features from various time-frequency representation. Features were extracted from a subset of 22 persons including 16 normal participants and 6 pathological patients. Accuracy was reported 98\% for this problem where the likelihood of over-training was used as parameters for KNN. The work investigated in~\cite{ref19} also chose KNN for clustering the samples into normal and pathological. This study also employed two approaches for dimensionality reduction of extracted time-frequency features; linear decomposition and tiling partition of mentioned features plane. Results were achieved on total of 45 recordings; including 19 pathological and 26 normal, and an average accuracy of 99\% was reported with 11-fold cross-validation.\par

\begin{table}[htbp]
  \centering
  \scriptsize
  \caption{Summary of the previous heart sound works, methods, database and results~\cite{ref2}.}
    \begin{tabular}{>{\centering\arraybackslash}m{80pt}>{\centering\arraybackslash}m{60pt}>{\centering\arraybackslash}m{60pt}p{15pt}p{15pt}p{15pt}p{25pt}}
    \hline
    Author & Database  & Method  & Se\%  & Sp\% & P+\% & Acc\%\\
    \hline
    \hline
    {\color{blue}Moukadem \it{et al.} (2013)}     & -    & Segmentation     & 96/97 & - & 95     & -\\
    \hline
    {\color{blue} Sun \it{et al.} (2014) } & -     & Segmentation     & -     & - & -     & 96.69\\
    \hline
	{\color{blue} Yan \it{et al.} (2010)}   & -     & Segmentation     & -     & - & -     & 99\\
    \hline
    {\color{blue}Sedighian \it{et al.} (2014)} & PASCAL     & Segmentation     & -     & - & - & 92.4/93.5\\
    \hline
    {\color{blue}Castro \it{et al.} (2013) }& PASCAL     & Segmentation     & -     & - & - & 90.9/93.3\\
    \hline
    {\color{blue}Schmidt \it{et al.} (2010)}   & -  & Segmentation     & 98.8  & - & 98.6     & -\\
    \hline
    {\color{blue}Sepehri \it{et al.} (2008)}     & 36 normal and 54 pathological & Frequency+ ANN     & 95  & 93.3    & - & -\\
    \hline
    {\color{blue} Uguz (2012)}    & 40 normal, 40 pulmonary and 40 mitral steno     & Time-frequency + ANN & 90.48 & 97.44 & - & 95\\
    \hline
    {\color{blue}Ari \it{et al.} (2010) } & 64 patients (normal and pathological)     & Wavelet + SVM     & -     & - & - & 86.72\\
    \hline
    {\color{blue}Zheng \it{et al.} (2015)}  & 40 normal and67 pathological     & Wavelet + SVM  & 98.8  & 99.3 & - & 98.9\\
    \hline
    {\color{blue}Gharehbaghi \it{et al.} (2015)} & 30 normal, 26 innocent and 30 AS     & Frequency + SVM  & 86.4  & 89.3 & - & -\\
    \hline
    {\color{blue}Saracoglu (2012)}  & 40 normal, 40 pulmonary and 40 mitral stenosis  & DFT and PCA + HMM  & 95    & 98.8 & - & 97.5\\
    \hline
    {\color{blue}Quiceno-Manrique \it{et al.} (2010)}    & 16 normal and 6 pathological     & Time-frequency + KNN     & - & - & - & 98 \\
    \hline
    {\color{blue}Avendano-Valencia \it{et al.} (2010)}    & 16 normal and 6 pathological     & Time-frequency + KNN & 99.56 & 98.54 & - & 99 \\
    \hline
    {\color{blue}Puri \it{et al.} (2016)}     & Physionet 2016     & mRMR + SVM & 77.49 & 78.91 & - & - \\
    \hline
    {\color{blue}Zabihi \it{et al.} (2016)}  & Physionet 2016     & Time-frequency + ANN & 85.9  & 86.91 & - & 84.9\\
    \hline
    {\color{blue}Potes \it{et al.} (2016)}     & Physionet 2016     & Time-frequency and AdaBoost + CNN  & 94.24 & 77.8 & - & - \\
    \hline
    {\color{blue}Rubin  \it{et al.} (2016)}    & Physionet 2016     &   MFCC + CNN   & 75    & 100 & - & 88\\
    \hline
    \end{tabular}%
  \label{tab1}
\end{table}

Table\ref{tab1} summarizes the research studies cited in this section.

Although the i-vector was originally used for speaker recognition applications~\cite{ref25}, it is currently used in various fields such as language identification~\cite{ref26,ref27}, accent identification~\cite{ref28}, gender recognition, age estimation, emotion recognition~\cite{ref29,ref30}, audio scene classification~\cite{ref31}, spoofing detection in automatic speaker verification systems~\cite{ref64}, etc. In this study, the i-vector is adopted for normal/abnormal heart sound classification task. Our approach is categorized in the time-frequency category, where MFCC is extracted and then the i-vector method is employed to extract features based on individuals' heart sound characteristics. \par

\section{Proposed Framework}
In this study, the proposed framework aims at using the i-vector for normal/abnormal heart sound classification. First, MFCC feature vectors are extracted from heart sound records. Then, In order to extract i-vector, a large GMM (e.g. 2048 components), called Universal Background Model (UBM) is trained using extracted MFCC features from all heart sound records (i.e. both normal and abnormal) in the training set. Subsequent to UBM training, zero and first-order statistics of the training features are extracted, accordingly. Then, these statistics are used to train the i-vector extractor through several iterations of the EM algorithm which will be explained in Section 3.2.4. After training the i-vector extractor, i-vectors are extracted from all records in the training set. At this stage, a fixed length i-vector is extracted for each record and then is fed into PCA or VAE in order to reduce its size and the intra-class variation as well. Eventually, there is a representative i-vector for each record, which will be used for classification.\par
\begin{figure}[!h]
\centering
\includegraphics[scale = 0.17]{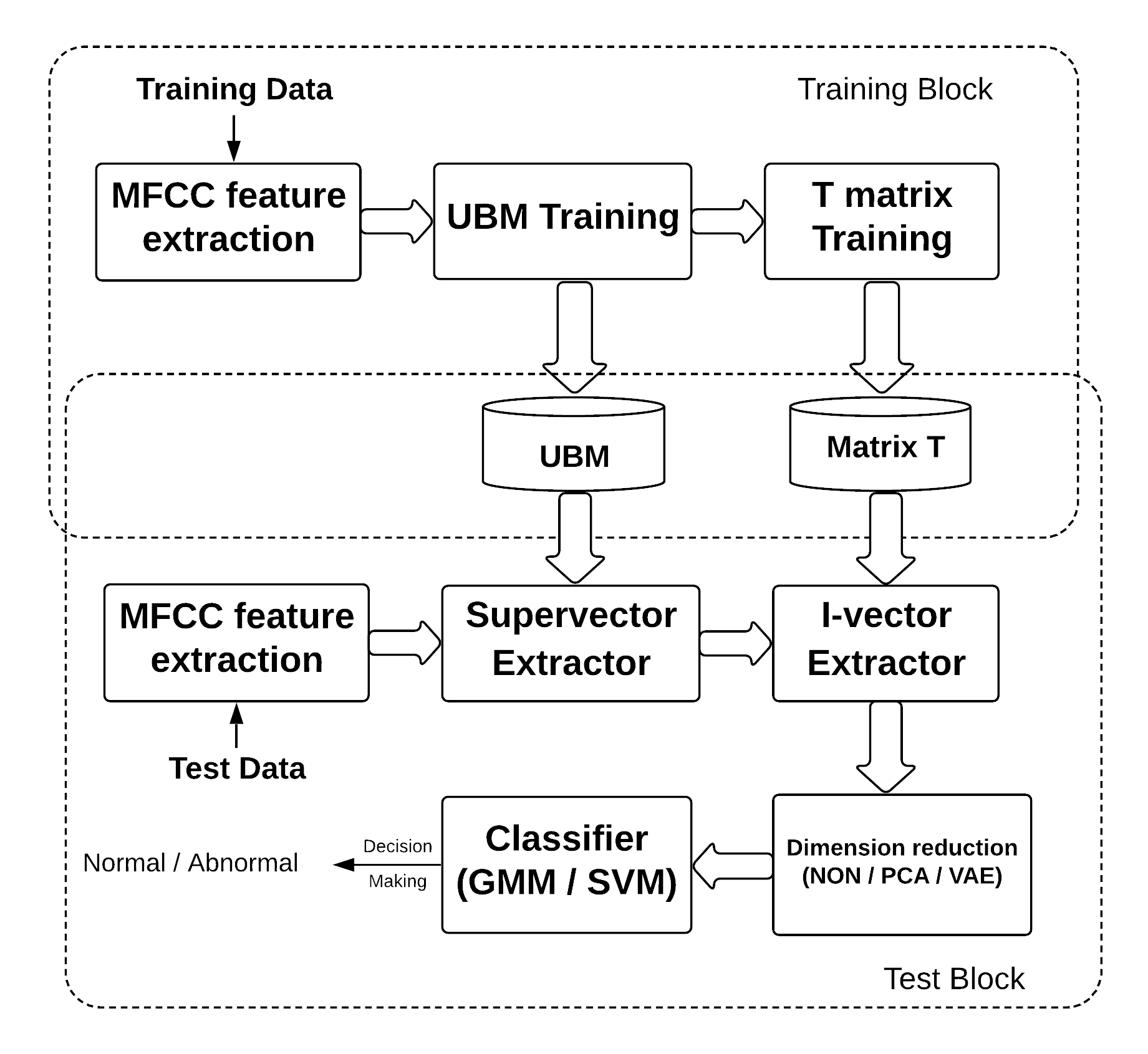}
\caption{Block diagram of the proposed system.}
\label{fig1}
\end{figure}

Fig.~\ref{fig1} briefly illustrates our proposed system.
\subsection{Mel-frequency Cepstral Coefficients}
MFCCs were employed over the years as one of the most salient features for speaker recognition~\cite{ref33}. The MFCC attempts to model human hearing perceptions by focusing on low frequencies (0-1KHz)~\cite{ref34}. In better words, the differences of critical bandwidth in the human ear are the basis of what we know as MFCCs. In addition, Mel frequency scale is applied to extract critical features of speech, especially its pitch.
\subsubsection{MFCC Extraction}
In the following section, we will explain how the MFCC feature is extracted. Initially, the given signal $s[n]$ is pre-emphasized. The concept of "pre-emphasis" means the reinforcement of high-frequency components passed by a high-pass filter~\cite{ref33}. The output of the filter is as follows
\begin{equation}\label{eq1}
p[n]=s[n]-0.97s[n-1].
\end{equation}
In the next step, named as framing, the pre-emphasized signal is dividing into same length short-time frames({\it e.g.} $25ms$) in order to achieve stationarity. Subsequently, the Hamming windows is applied as

\begin{equation}\label{eq2}
h[n] =p[n] \times \left\{ 0.54-0.46 \cos \left(\frac{2\pi n}{N-1}\right) \right\} \quad 0<n<N-1,
\end{equation}
where $N$ is the number of samples in each frame. Heart sounds are sampled by 2 KHz frequency ratio, each frame has length of 25 $ms$, and number of samples is 50. \par
To analyze $h[n]$ in the frequency domain, an N-point Fast Fourier Transform (FFT) is applied to convert it into the frequency domain according to
\begin{equation}\label{eq3}
H[k] =\sum_{n=0}^{N-1}{h[n] e^{-j\frac{2k\pi n}{N}}}.
\end{equation}
A logarithmic power spectrum is obtained by log energy computation block, on a Mel-scale using a filter bank that consists of $L$ filters.
\begin{equation}\label{eq4}
X[l]=\log\left(\sum_{k=k_{ll}}^{k_{lu}}{\left| H[k] \right| W_{l}\left(k\right)}\right)\quad l=0, 1, \dots, L-1,
\end{equation}
where $H_l(k)$ is the absolute value of complex Fourier transform, $W_l(k)$ is the $l$th triangular filter, $k_{ll}$ and $k_{lu}$ are the lower limit and upper limit of the $l$th filter, respectively. In our experiment, the number of filters in the filter bank; $L$, was set to 20.\par
\begin{figure}[t]
\centering
\includegraphics[scale = 0.48]{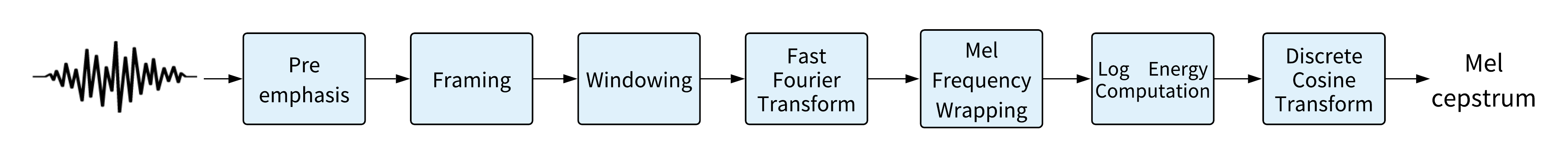}
\caption{Block diagram of MFCC feature extraction~\cite{ref26}.}
\label{fig2}
\end{figure}

The given frequency $f$ in hertz can be converted to Mel-scale as follow
\begin{equation}\label{eq5}
F\left(Mel\right)=2595 \times \log 10\left(1+f/700\right).
\end{equation}
Eventually, the MFCCs coefficients are obtained by applying Discrete Cosine Transform (DCT) to the $X[l]$
\begin{equation}\label{eq6}
C[m] = \sum_{l=1}^{L}{X[l] \cos \left[ \frac{\pi m\left(l-0.5\right)}{L}\right]} \quad m=1, \dots , M-1.
\end{equation}
where m is the index of obtained MFCC components and $M$ is the number of MFCC features, which was set to 12. The steps for extracting the MFCC features are depicted in Fig.~\ref{fig2}.  
\subsection{i-Vector}
The aim of this paper is to propose a framework for the heart sound classification using i-vector which was first proposed for speaker recognition application~\cite{ref25} and later was adopted in other applications such as language identification~\cite{ref26}, emotion recognition~\cite{el2011survey}, music genre classification~\cite{wei2017multilingual}, and online signature verification~\cite{ref40}, etc. i-vector can be considered as a technique to map a sequence of feature vectors for a given sample into a low-dimensional vector space, referred to as the total variability space, based on a factor analysis technique. In other words, it is a technique to extract a compact fixed-length representation given a sequence of feature vectors with arbitrary length. Then, the extracted compact feature vector can be either used for vector distance-based similarity measuring or as input to any further feature transform or modelling.

There are determined steps to extract i-vector from a heart sound record. First, MFCC feature vectors should be extracted from the input signal and then the Baum–Welch statistics should be extracted from the features, and finally i-vector is computed using these statistics. In the following subsections, we go through these steps in details.

\subsubsection{Universal Background Model Training}
The first step in implementing i-vector extraction pipeline is to create a global model which is called UBM which is used to map the features to a high-dimensional space to give a better representation. Gaussian mixture models (GMMs) have been frequently used for building an UBM, especially in the text-independent speaker verification task~\cite{ref25,ref39}. GMM estimates the distribution of extracted MFCC features using of a finite number of Gaussian distributions.  Here, the GMM model is trained by MFCC features from all heart sound records in the training set which is supposed to be large enough to cover all the feature space.

\subsubsection{Extraction of Baum–Welch Statistics}
Here, for each MFCC feature sequence, the zero and first-order Baum-Welch statistics are extracted using UBM which is modeled by a GMM. ~\cite{ref38,ref45}.

Suppose $X_i$ as the whole feature vectors collected to train $i$th heart sound; then the zero and first-order statistics for the $c$th component of UBM named $N_c$ and $F_c$ are calculated as follows:
\begin{equation}\label{eq8}
N_{c}\left(X_{i}\right)= \sum_{t}{\gamma _{i,t}^{c}}
\end{equation}
\begin{equation}\label{eq9}
F_{c}\left(X_{i}\right)= \sum_{t}{\gamma_{i,t}^{c}}\left(X_{i,t}-m_{c}\right),
\end{equation}

where $X_{i,t}$ is the $t$th MFCC feature vector for heart sound $i$th, $m_c$ indicates mean of $c$th component, and finally $\gamma_{i,t}^{c}$ shows the posterior probability of $X_{i,t}$ by the $c$th component described as below:
\begin{equation}\label{eq10}
\gamma _{i,t}^{c}=Pr\left(c| X_{i,t}\right)=\frac{w_{c}N\left(X_{i,t}|m_{c},\Sigma_{c}\right)}{\sum_{j=1}^{C}{w_{j} N\left(X_{i,t}|m_{j},\Sigma_{j}\right)}}.
\end{equation}

\subsubsection{i-vector Extraction}
Suppose $M$ is a mean-supervector which represents the feature vectors of a heart sound record. Supervector of each record is a $DC$-dimensional vector obtained by concatenating $D$-dimensional mean vectors of the its GMM. GMM for each record is obtained by MAP adaptation. The supervector of $i^{th}$ record is modelled as follows ~\cite{ref25}:

\begin{equation}\label{eq11}
M_i=m+Tw_i,
\end{equation}
where $m$ is an independent mean-supervector ($m = [m_1^t, m_2^t,..., m_C^t]^t$) extracted from the UBM, $T$ is a low-rank matrix, and $w_i$ represents a random latent variable with a standard normal distribution for $i^{th}$ record. $M_i$ is assumed to has a Gaussian distribution with mean $m$ and covariance matrix $TT^T$, where $T^T$ is regarded as transpose of $T$. The i-vector is the MAP point estimation of the variable $w_i$ which is equal to the mean of the posterior probability of $w_i$ given the $i^{th}$ record.

In Eq.~\ref{eq11}, $m$ and $T$ as parameters should be estimated. $m$ as a mean-supervector is obtained by concatenating the means of the UBM components ~\cite{ref45}. To obtain $T$, expectation maximization (EM) is applied. Assume the UBM has $C$ components ( in this work, is set to 2048), and dimensions of feature vectors are $D$, the matrix $\sum$ is described as

\begin{equation}\label{eq12}
\Sigma=
    \begin{pmatrix}
      \Sigma_1 & 0 & ... & 0 \\
      0 & \Sigma_2 & ... & 0 \\
      . & . & ... & 0 \\
      0 & 0 & ... & \Sigma_C
    \end{pmatrix},
\end{equation}

where $\Sigma_{c}$ is the covariance matrix of the $c^{th}$ component of of the UBM. Let $X_i$ be all feature vectors of $i^{th}$ record and $P\left(X_i|M_i,\Sigma\right)$ indicates the likelihood of $X_i$ computed with the GMM specified by the supervector $M_i$ and the super-covariance matrix $\Sigma$, then the EM optimization can be performed by iterating the following two steps. First, the current value of matrix $T$ is used to estimate the vector that maximize the likelihood as follows:
\begin{equation}\label{eq13}
w_i=arg\max_{w}{P\left(X_i|m+Tw,\Sigma\right)}.
\end{equation}
Then, $T$ is updated by maximizing the following relation:
\begin{equation}\label{eq14}
\prod_{i}{P(X_{i}|m+Tw_{i},\Sigma)}.
\end{equation}
By taking the logarithm of Eq.~\ref{eq14}, log-likelihood of each record can be computed as:
\begin{equation}\
\label{eq15}
\begin{split}
   \log \left(P(X_{i}|m+Tw_{i},\Sigma)\right) & = \sum_{c}N_c\log{\frac{1}{\left(2\pi\right)^{D/2}\left|\Sigma_c\right|^{1/2}}} \\
     & -\frac{1}{2}
    \sum_t{\left(X_{i,t}-T_cw_i-m_c\right)^t \Sigma_{c}^{-1}{\left(X_{i,t}-T_cw_i-m_c\right)}},
\end{split}
\end{equation}
where $c$ iterates over all components of the UBM and t iterates over all feature vectors and $T_c$ is a submatrix of $T$ which is related to the $c^{th}$ component. Let the zero and the first-order statistics have been calculated by Eq.~\ref{eq8} and Eq.~\ref{eq9}, respectively, the the posterior covariance matrix, $Cov(w_i,w_i)$, mean $E[w_i]$, and the second moment $E[w_i w_i^t]$ are computed for $w_i$ as:
\begin{equation}\label{eq16}
Cov\left(w_i,w_i\right)=\left(I+\sum_{c}N_c\left(X_i\right)T_c^t\Sigma_c^{-1}T_c\right)^{-1}
\end{equation}
\begin{equation}\label{eq17}
E\left[w_i\right]=Cov\left(w_i,w_i\right)\sum_{c}T_c^t\Sigma_c^{-1}F_c\left(X_i\right)
\end{equation}
\begin{equation}\label{eq18}
E\left[w_iw_i^t\right]=Cov\left(w_i,w_i\right)+E\left[w_i\right]{E\left[w_i\right]}^t.
\end{equation}
Ultimately, by maximizing Eq.~\ref{eq14}, the updated value of T can be calculated as 

\begin{equation}\label{eq19}
T_c = \left(\sum_{i}^{}F_c(X_i)E[w_i]^t\right)\left(\sum_{i}^{}N_c(X_i)E[w_iw_{i}^{t}]\right)^{-1}.
\end{equation}

As said before, i-vector is the mean of the posterior probability of $w_i$ given $i^{th}$ input record where $w_i$ is a random hidden variable with a standard normal distribution. To extract i-vector, the MAP point for $w$ is estimated and it formula is described as Eq.\ref{eq17}.

\subsection{Techniques for reducing the feature dimension and the effects of intra-class variations}
There are several techniques for reducing the feature dimension and the effects of intra-class variations. In the i-vector based applications, various techniques such as nuisance attribute projection (NAP)~\cite{ref25,ref45,ref46,ref47}, within-class covariance normalization (WCCN)~\cite{ref25,ref48,ref49}, principal component analysis (PCA)~\cite{ref49}, and linear discriminant analysis (LDA)~\cite{ref50} are extensively employed. In this work, PCA and a new emerging technique called Variational Autoencoders (VAE)~\cite{ref51} are employed which will be explained in the following subsections.
\subsubsection{Principal Component Analysis}
In this method, important information is extracted from the data as new orthogonal variables, which are referred to as the principal components~\cite{ref52}. To achieve this objective, assume a given $n\times p$ zero mean data matrix $X$ where $n$ and $p$ indicate the number of feature vectors and feature size, respectively. Accordingly, to define the PCA transformation consider vector $x_{(i)}$ of $X$ which is mapped by a set of p-dimensional vectors of weights $w_{(k)}={(w_1,\dots,w_p)}_{(k)}$ to a new vector of principal component  $t_{(i)}={(t_1,\dots,t_l)}_{(i)}$, as follows
\begin{equation}\label{eq20}
t_{k_{(i)}}= x_{(i)}\; .\; w_{(k)}; \quad i=1,\ldots ,n\quad k=1,\ldots ,l,
\end{equation}
where vector $t$ (consists $t_1,\ldots,t_l$) inherits the maximum variance from $x$ by weight vector $w$ constrained to be a unit vector~\cite{ref53}.
\subsubsection{Variational Autoencoder}
\begin{figure}[t]
\centering
\includegraphics[scale = 0.20]{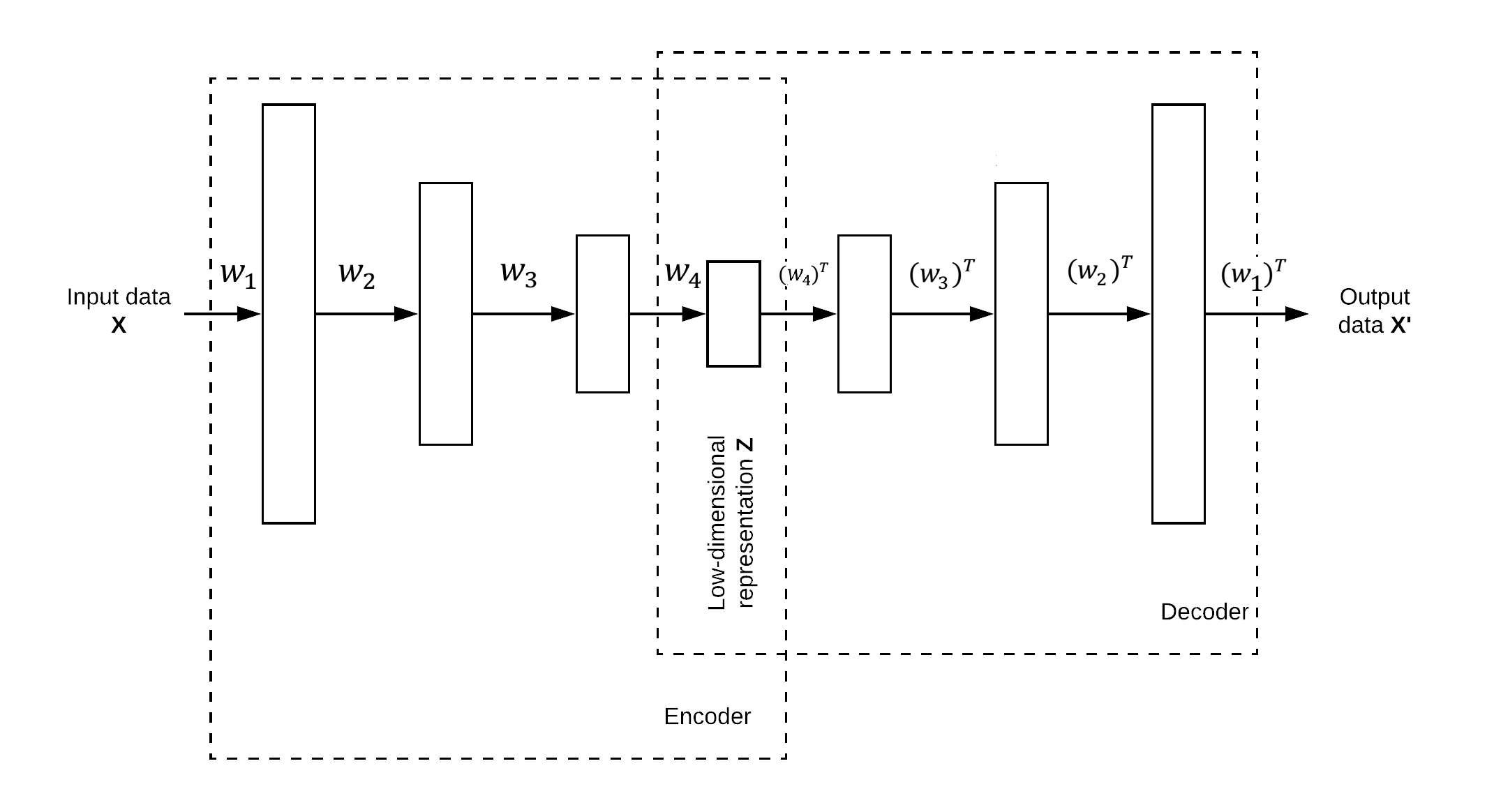}
\caption{Block diagram of VAE.}
\label{fig3}
\end{figure}

As one of the most prominent approaches to extract valuable information is VAE which is among the generative models. This model attempts to reconstruct data from input data. In this regard, Consider $x$ as the input for a VAE which seeks to encode the inputs into latent variables $z$, and then reconstructed input $x'$ will be produced from the latent variables. To this end, the training process aims to minimize the cost function (Mean Square Error (MSE) between input and output). In the optimal situation, the input and output are the same. The architecture of VAEs comprise hidden layers~\cite{ref51} with odd numbers and $d$ nodes. The weights are shared between top and bottom layers, which both have $D$ nodes. Schematic of a VAE is depicted in Fig.~\ref{fig3}.
AS shown in Fig.~\ref{fig3}, encoded variable $z$ can be used as enhanced features for the better description of input $x$. To obtain the vector $z$, a probability function on $x$, called $p(x)$, is defined, seeking to maximize likelihood of the mentioned probability; $\log p(x)$~\cite{ref54}. $E_{\left(z\sim q(z|x)\right)}$ shows the expectation of random variable $z$ over probability function $q(z|x)$. Since we have no information about $p(z|x)$; an approximation of $p(z|x)$, called $q(z|x)$, is computed. Thus, based on Bayes rule we have~\cite{ref54}
\begin{equation}\label{eq21}
\log p(x)=E_{\left(z\sim q(z|x)\right)}\left[\log p(x)\right]=E_{\left(z\sim q(z|x)\right)}\left[\log \frac{p(x|z)p(z)}{p(z|x)}\right],
\end{equation}
here, we multiply and divide the term by $q(z|x)$ as an approximation for $p(z|x)$
\begin{equation}\label{eq22}
E_{\left(z\sim q(z|x)\right)}\left[\log\frac{p(x|z)p(z)q(z|x)}{p(z|x)q(z|x)}\right].
\end{equation}
So, it can be concluded that
\begin{equation}\label{eq23}
E_{\left(z\sim q(z|x)\right)}\left[\log p(x|z)\right]+E_{\left(z\sim q(z|x)\right)}\left[\log \frac{p(z)}{q(z|x)}\right]+ E_{\left(z\sim q(z|x)\right)}\left[\log \frac{q(z|x)}{p(z|x)}\right].
\end{equation}
Finally
\begin{equation}\label{eq24}
\underbrace{E_{\left(z\sim q(z|x)\right)}\left[\log p(x|z)\right]-E_{\left(z\sim q(z|x)\right)}\left[\log \frac{q(z|x)}{p(z)}\right]}_{A}+ \underbrace{E_{\left(z\sim q(z|x)\right)}\left[\log \frac{q(z|x))}{p(z|x)}\right]}_{B},
\end{equation}
the term $B$ is intractable, and has a value greater than zero. As a result, the term $A$ is attempted to be minimized as a tractable lower bound. The log-likelihood measure is a good indicator to show how much samples from $q(z|x)$ can describe data $x$.\par

It is noteworthy that VAEs are a good solution for different problems such as missing data imputation and so forth~\cite{ref51}.

\subsection{Gaussian Mixture Models}
In this study, GMMs are engaged as a classifier for the extracted features from heart sound records. GMMs are among models with the probabilistic nature, which are suitable for general distributions consisted of sub-populations  ~\cite{ref650}. GMMs use an iterative process to determine which data point belongs to each sub-population, without any knowledge about data point labels. Hence, GMMs are considered as unsupervised learning models.

The GMM is introduced with two types of parameters: the weights of the Gaussian mixture components and the means and the variance of the Gaussian mixture components. The Probability Distribution Function (PDF) of a $K$ components GMM, with mean $\mu_{k}$ and covariance matrix $\Sigma_k$ for the $k^{th}$ component is defined as
\begin{equation}\label{eq25}
p({x})=
\sum_{i=1}^{K}{\frac{1}{\sqrt{(2\pi )^{K}\left| \Sigma_{i}\right| }}\phi
_{i}exp\left(-\frac{1}{2}({x}-{\mu
_{i}})^{T}\Sigma_{i}^{-1}({x}-{\mu _{i}})\right)}
\end{equation}
\begin{equation}\label{eq26}
\Sigma_{i=1}^{k}{\phi _{i}=1},
\end{equation}
where ${x}$ is a feature vector and $\phi_i$ is the weight of the mixture component $i^{th}$.

If the number of components is defined, Expectation Maximization (EM) is a method that is often used to estimate the parameters of the mixture model. In frequentist probability theory, models are usually learned using maximum likelihood estimation techniques. The maximum probability estimate is engaged to maximize the probability or similarity of the observed data with respect to the model parameters~\cite{ref55}.
The maximization of EM is a numerical method for estimating the maximum probability. Maximization of EM is a repetitive algorithm and has the property that the most similarity of data with each subsequent replication increases significantly, which means that it achieves to the maximum point or the local maximum point~\cite{ref55}.

The maximizing likelihood estimation of Gaussian mixture models includes two steps. The first step is known as "expectation", which includes calculating the expectation and assigning the $k^{th}$ component ($C_k$) for each $x_i\in X$ data point with the parameters of the model $\phi_k$, ${{\mu}}_{k}$ and $\Sigma_k$. The second step is known as "maximization", which includes maximizing the expectation calculated in the previous step relative to the model parameters. This step involves updating the values of $\phi_k$, ${\mu}_{k}$ and $\Sigma_k$.\ The entire process is repeated as long as the algorithm converges, giving maximum likelihood estimation. More details are available at~\cite{ref55}.

\subsection{Support Vector Machine}
We also employed SVM as a different classifier to compare the obtained results with those of the GMM. Thus, a brief overview of this classifier is presented below. 
In this method, a hyperline is used to discriminate between samples by returning a solution to a two-class classification problem:
\begin{equation}
    \label{eq:svm}
    y(x) = w^T\phi(x) + b,
\end{equation}
where, $w$ is an unknown weight matrix to learn. $\phi(x)$ denotes a fixed feature-space transformation and $b$ is the bias parameter. Consider $N$ training data $x_1, ..., x_N$ with target values $t_1, ..., t_N$ ($t_n \in \{-1,1\}$). Each data is classified based on sign of $y(x)$. Therefore, both $t_n$ and $y(x_n)$ should have the same sign, and $t_n y(x_n)>0$. There will be multiple solutions for each SVM problem, but the one with the smallest generalization error is desirable. "Margin" is the term to describe the smallest distance between the decision boundary and any of the samples. The solution with the maximum value of the margin is chosen as the best solution. 
Considering margin definition, the distance from a point $x_n$ to the decision surface is calculated by:
\begin{equation}
    \frac{t_n y(x_n)}{||w||} = \frac{t_n (w^T\phi(x_n)+b)}{||w||}.
\end{equation}
We seek to optimize the parameters $w$ and $b$ to maximize the distance. This can be achieved by:
\begin{equation}
    \arg \max_{w,b} \{\frac{1}{||w||}\min_n [t_n(w^T\phi(x_n)+b)]\}
    \label{eq:svm-optimization}.
\end{equation}
This problem can be converted to a less complex problem for easier solving. Since the scale has no effect on the solution there is freedom to consider the relation in Eq. \ref{eq:svm-optimization} as follows:
\begin{equation}
    t_n(w^T\phi(x_n)+b) = 1.
\end{equation}
The optimization problem in Eq. \ref{eq:svm-optimization} requires to maximize $||w||^{-1}$, equal to minimizing $||w||^{2}$. So we also have to solve the following optimization problem:
\begin{equation}
    \label{eq:svm-constraint}
    \arg \min_{w,b}\frac{1}{2}||w||^2.
\end{equation}
To solve Eq. \ref{eq:svm-constraint} , Lagrange multipliers $a_n \geq 0$ is introduced for each constraint. So the Lagrange equation would be like:
\begin{equation}
    L(w,b,a) = \frac{1}{2}||w||^2 - \sum_{n=1}^{N}a_n\{t_n(w^T\phi(x_n)+b)-1\}.
    \label{svm-lagrange}
\end{equation}
Then derivatives of $L(w,b,a)$ are set to zero with respect to $w$ and $b$. Then:
\begin{equation}
\label{eq:svm-w}
    w = \sum_{n=1}^{N}a_n t_n\phi(x_n)
\end{equation}
\begin{equation}
    \sum_{n=1}^{N}a_n t_n  = 0.
\end{equation}
Considering Eq. \ref{eq:svm-w},  the Eq. \ref{eq:svm}  can be rewritten as follows:
\begin{equation}
\label{eq:svm-rewritten}
    y(x) = (\sum_{n=1}^{N}a_nt_n\phi(x_n)\phi(x)) + b.
\end{equation}
The form of Eq. \ref{eq:svm-rewritten} allows the model to be reformulated with kernels. Thus, a solution for problems with infinite feature space can be obtained by:
\begin{equation}
\label{eq:svm-kernel}
    y(x) = (\sum_{n=1}^{N}a_nt_nk(x_n,x)) + b,
\end{equation}
where $k(x_n,x)$ is a positive definite kernel.\par
For this work Radial Basis Function (RBF) was used as kernel for SVM. The formulaiton of this kernel is given below:
\begin{equation}
    K(x,x') = \exp(-\frac{||x-x'||^2}{2\sigma^2}).
\end{equation}

\section{Experimental Setup}
\subsection{Dataset}
The 2016 Physionet/CinC challenge is introduced to provide a standard database containing normal and abnormal heart sound~\cite{ref2}. The presented dataset in this challenge is a heart sound recording set of subjects/patients, collected from a variety of environmental conditions (including noisy conditions with low signal quality) as described in~\cite{ref2}. Therefore, the majority of heart sounds have incurred different noises during recordings such as speech, stethoscope motion, breathing and intestinal activity~\cite{ref2}. These noises complicate the classification of normal and abnormal heart sounds. Accordingly, the organizers allowed the participants to classify some of the recordings as 'unsure'~\cite{ref2} and it indicates the difficulty level of the challenge. This dataset consists of three subsets: training, validation, and test. For training purposes, six labeled databases (names with the prefix $a$ to $f$) contain 3153 sound recordings from 764 subjects/patients, with the duration of 5-120 s).\par
The validation subset is comprised of 150 normal and 151 abnormal heart sound (with file names prefixed alphabetically, $a$ through $e$) and the test data includes 1277 heart sound trials generated from 308 subjects/patients. It has to be noted that 301 selected recordings from train set were used as a test set for validation.\par
The Challenge test set consisted of six databases labeled from b to $e$, $g$, and $i$ with 1277 heart sound recordings from 308 participants. The statistics of each subset are summarized and illustrated in Table 2. More details about the dataset and the 2016 Physionet/CinC challenge can be found in~\cite{ref2}.\par
In this study, the results were reported based on the publicly available part of the Physionet/CinC 2016 dataset. It is worth mentioning that the training set is divided into two parts via five phases. In each phase, we randomly assigned 80\% of the training set as our training set and the rest of 20\% was assigned as our validation set which is used for tuning the parameters. In addition, we used Physionet/CinC 2016 validation set as our test set. Details of the dataset are presented in Table \ref{tab2}.
\begin{table}[htbp]
  \centering
  \scriptsize
  \caption{Statistics of the 2016 Physionet/CinC dataset~\cite{ref2}.}
    \begin{tabular}{|c|>{\centering}m{40pt}|c|ccc|>{\centering}m{20pt}>{\centering}m{20pt}>{\centering}m{20pt}m{20pt}|}
    \hline
    Subset & \#Patients  & \#Records  & \multicolumn{3}{c|}{\#Proportion of recordings}  & \multicolumn{4}{c|}{\#The weight parameters}\\
     & & & Abnormal & Normal & Unsure & wa$_1$ & wa$_2$ & wn$_1$ & wn$_2$ \\
    \hline
    Training & 746 & 3153 & \multicolumn{1}{c|}{18.1} & \multicolumn{1}{c|}{73.03} & \multicolumn{1}{c|}{8.8} &
    \multicolumn{1}{m{20pt}|}{0.8602} & \multicolumn{1}{m{20pt}|}{0.1398} & \multicolumn{1}{m{20pt}|}{0.9252} & \multicolumn{1}{m{20pt}|}{0.0748} \\
    Eval. & - & 301 & \multicolumn{1}{c|}{-} & \multicolumn{1}{c|}{-} & \multicolumn{1}{c|}{-} &
    \multicolumn{1}{m{20pt}|}{0.7888} & \multicolumn{1}{m{20pt}|}{0.2119} & \multicolumn{1}{m{20pt}|}{0.9467} & \multicolumn{1}{m{20pt}|}{0.0533} \\
    Test & 308 & 1277 & \multicolumn{1}{c|}{12.1} & \multicolumn{1}{c|}{77.1} & \multicolumn{1}{c|}{10.9} &
    \multicolumn{1}{>{\centering}m{20pt}|}{-} & \multicolumn{1}{>{\centering}m{20pt}|}{-} & \multicolumn{1}{>{\centering}m{20pt}|}{-} & \multicolumn{1}{>{\centering}m{20pt}|}{-} \\
    \hline
    \end{tabular}%
  \label{tab2}
\end{table}

\subsection{Evaluation Metrics}
In this task, the metric of evaluation is based on Modified Accuracy (MAcc) introduced by Physionet 2016 challenge. For MAcc computation, data is categorized into three categories; normal, abnormal or unsure, with two references in each category. The modified sensitivity ($Se$) and specificity ($Sp$) can be computed according to:
\begin{equation}\label{eq30}
S_{e}= \frac{wa_{1 }\times
Aa_{1}}{Aa_{1}+Aq_{1}+An_{1}}+ \frac{wa_{2 }\times
(Aa_{2}+Aq_{2})}{Aa_{2}+Aq_{2}+An_{2}}
\end{equation}
\begin{equation}\label{eq31}
S_{p}= \frac{wn_{1 }\times
Na_{1}}{Na_{1}+Nq_{1}+Nn_{1}}+ \frac{wn_{2 }\times
(Na_{2}+Nq_{2})}{Na_{2}+Nq_{2}+Nn_{2}},
\end{equation}
where $wa_1$ and $wa_2$ are the percentages of the abnormal recordings of the signal with good quality and poor quality respectively, and  $wn_1$ and $wn_2$ are of the normal recordings of the signal with good quality and poor quality respectively. For all 3153 training set recordings, values for weight parameters of $wa_1$, $wa_2$, $wn_1$, and $wn_2$ are equal to 0.8602, 0.1398, 0.9252 and 0.0748 respectively, in the train set. These parameters were also calculated for validation set and were reported 0.78881, 0.2119, 0.9467 and 0.0533 respectively. The “Score” for this challenge is computed using the following equation
\begin{equation}\label{eq32}
MAcc=(S_{p}+S_{e})/2.
\end{equation}
\subsection{Scoring and Decision Making}
To assign a score to a given heart sound based on the GMM classifier we proceed as follows. First, an i-vector is extracted from the training set and is projected to the new space using the PCA or VAE. Afterwars, they are applied to two GMMs (one GMM for the normal heart sound and the other for the abnormal heart sound) with different components to learn the model by EM iterations (training GMMs). In the next step, the score for each trial is obtained by computing log likelihood ratio:
\begin{equation}\label{eq33}
LLR(S)=\log P(S|\theta _{normal})-\log P(S|\theta_{abnormal}),
\end{equation}
where $S$ is an i-vector corresponding to the test record, while $\theta_{normal}$ and $\theta_{abnormal}$ denote the GMMs for normal and abnormal heart sounds, respectively. Once the score is found, a simple global threshold is applied to it to make the final decision of normal/abnormal heart sound classification. If the score is higher than the threshold, the test heart sound is labeled as normal and otherwise, it is labeled as abnormal. In this study, a global threshold able to plot the detection error trade-off (DET) and detection accuracy trade-off (DAT) curves was used.
\section{Experimental Results}
In this section, first we briefly introduce the baseline system and in the following, to verify the performance of proposed framework, we carried out various experiments using the physionet 2016 dataset. In Section 5.2, we investigate the effect of GMM components and i-vector dimensionality. Ultimately, the effect using different size of training set is examined, in Section 5.3.

\subsection{Baseline System}
In this research study, the proposed approach in~\cite{ref56} is considered as the baseline system. The Physionet 2016 dataset is used in the baseline system in the same manner that we used in the proposed system. The proposed method in the baseline system is based on Mel-Spectrogram, MFCC and sub-band envelopes features and different configurations of CNN classifier. Accordingly, 103228 frames were extracted from the Physionet 2016 dataset. To report the results, they repeated their experiments in five iterations and reported the average of the obtained results. The attained results in terms of sensitivity, specificity and mean accuracy denoted 0.845, 0.785 and 0.815, respectively.

\subsection{Effects of the Number of the GMM Components and i-vectors Dimensionality}
\begin{figure}[H]
\centering
\includegraphics[scale = 0.45]{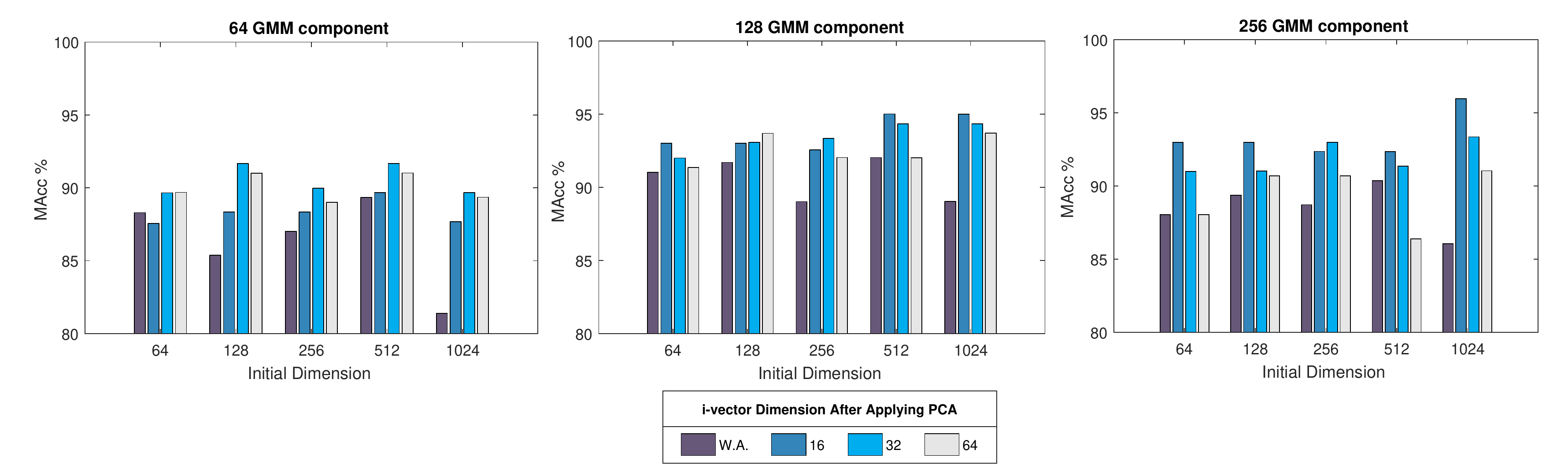}
\caption{MAcc comparison based on different GMM components, raw i-vector dimension and dimension of i-vector using "PCA" for the proposed method. Here the Acronym "W.A" means "Without Applying" PCA.} 
\label{fig:PCA}
\end{figure}

\begin{figure}[H]
\centering
\includegraphics[scale = 0.45]{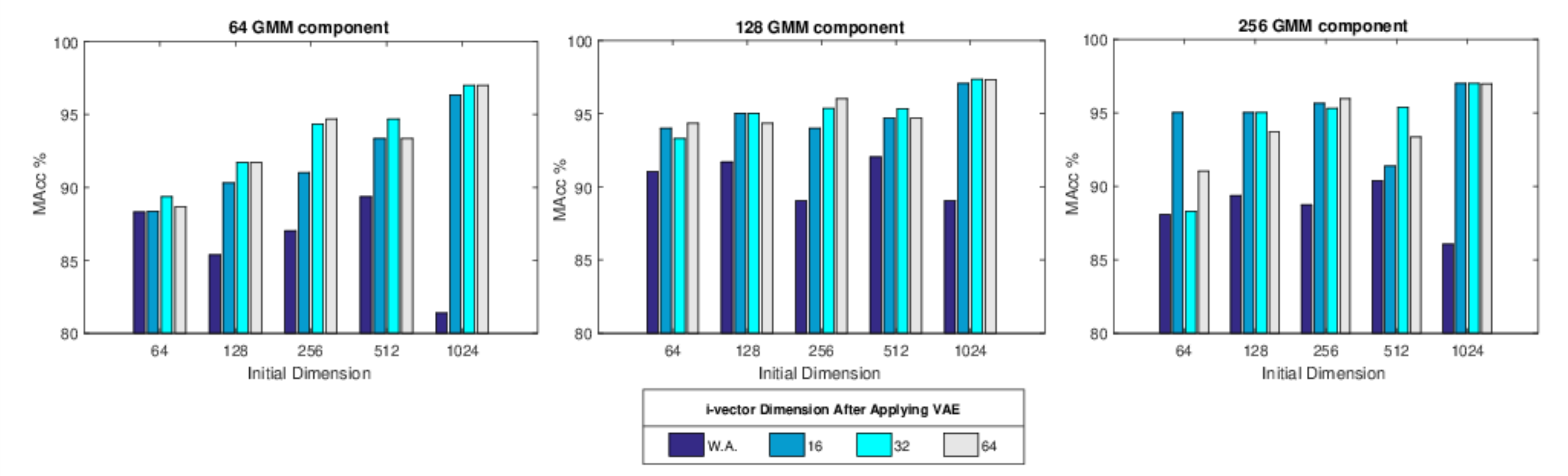}
\caption{MAcc comparison based on different GMM components, raw i-vector dimension and dimension of i-vector using "VAE" for the proposed method. Here the Acronym "W.A" means "Without Applying" VAE.} 
\label{fig:VAE}
\end{figure}

The first part of our experiments was performed to investigate the effects of the number of GMM components, the effects of i-vectors dimension numbers without applying VAE or PCA and finally, effects of i-vector dimension reduction through applying PCA and VAE on the proposed system. Fig.~\ref{fig:PCA} and Fig.~\ref{fig:VAE} represent the MAccs on the test set using the mentioned approaches.  It is worth mentioning that we did not label any data as “unsure”, and the label “normal” or “abnormal” is assigned to all test data.\par

In Fig.~\ref{fig:PCA} and ~\ref{fig:VAE} the number of components used in GMMs is specified separately in each plot. Fig.~\ref{fig:PCA} and Fig.~\ref{fig:VAE} show i-vector generally performs better after the application of VAE or PCA. The best results are achieved by higher dimensions of i-vector and after the application of VAE. Fig.~\ref{fig:PCA} denotes the results of i-vector and its PCA. It can be seen that the values obtained after employing PCA are not as good as those obtained by employing VAE.\par
\textbf{Discussion:} The higher performance of VAE is due to the fact that it aims to minimize the cost function which is defined as MSE between input features and output (reconstructed features). PCA merely seeks to extract important information, whereas VAE attempts to extract features with the capability to produce original data. As a result, VAE can extract valuable information which is able to produce original data as much as it can and that is why MAcc is reducing over time. On the other hand, increase in dimension of raw i-vector might add useless, sparse features to the feature vector and this leads to classification error and accuracy reduction.
Generally, the best MAcc values are obtained by the GMMs trained by 128 components. In the proposed system, the GMMs are not well trained with 64 components. Conversely, engaging 256 components cause over-fitting, due to the low amount of training set.\par

\subsection{Effects of i-vector Dimension on Performance}
\begin{figure}[H]
\centering
\includegraphics[scale = 0.5]{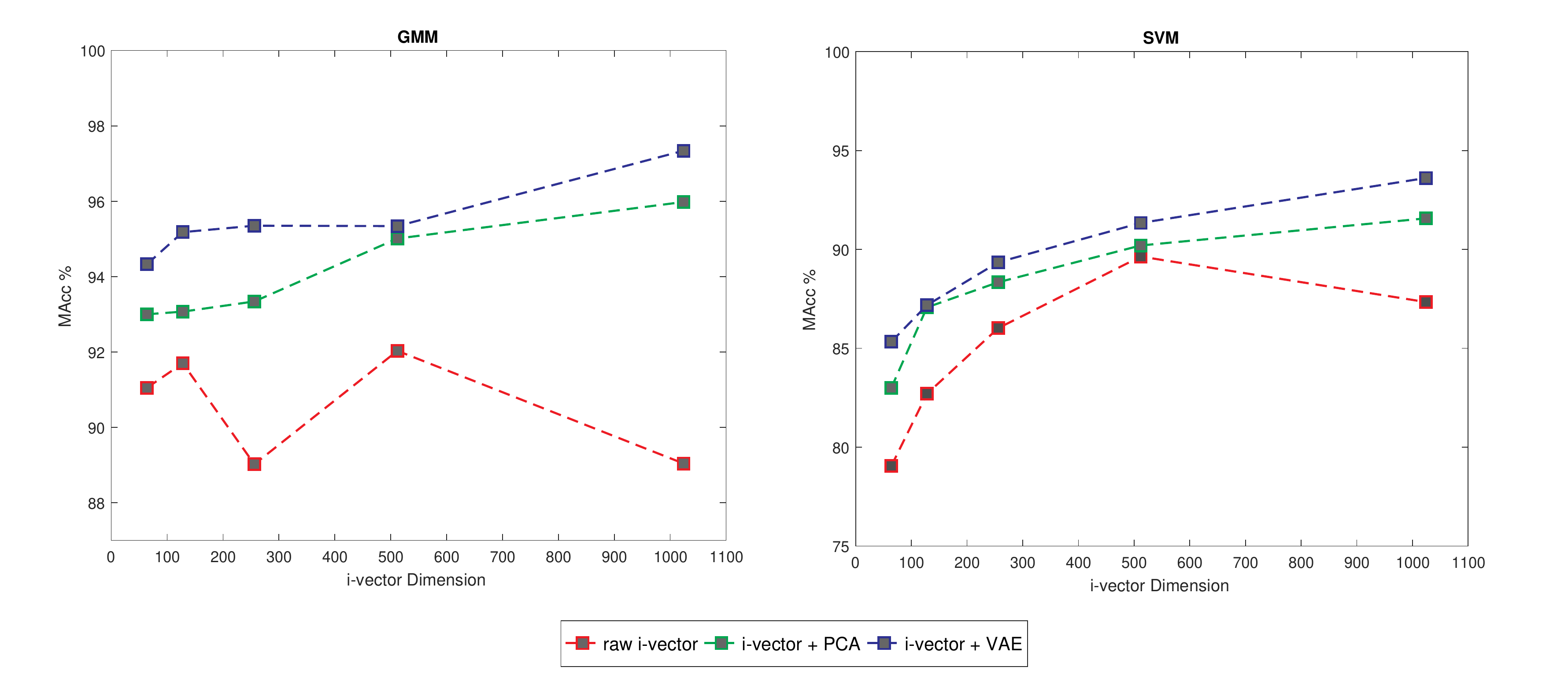}
\caption{DAT curve comparison for raw i-vevtor and its PCA and VAE. In each case, results are reported using the best parameters configuration.}
\label{fig5}
\end{figure}

In Fig.~\ref{fig5} The red point-line represents the best values achieved by different dimensions of i-vectors without applying PCA or VAE. Moreover, the blue and green point-lines of Fig.~\ref{fig5} represent the best values obtained by different dimensions of i-vectors and applying PCA and VAE, respectively. According to the
Fig.~\ref{fig5}, after employing VAE or PCA, the MAcc values subsequently increase. However, this pattern is not true for raw i-vectors which yield different MAcc results.\par
\textbf{Discussion:} A higher-dimensional i-vector includes more detailed information. On the other hand, this information may include useless details and common information. Therefore, PCA and VAE methods are adopted to make this information more effective. Applying PCA and VAE can significantly improve the result values compared with applying raw i-vectors. In addition, it can be seen that although GMM works better for higher dimensions, SVM has better improvement rate than GMM. This happens owing to the fact that SVM is a method which works based on feature space transformation; therefore, any change in the dimensions of features can be more effective than GMM, which is more data-based classifier. It will be demonstrated in Section 5.4 that GMM displays a better improvement rate when the data is increased, while SVM has a smooth improvement curve.  

\subsection{Effect of Training the System using Different Size of Training Set}
The section is concerned with evaluating the effect of different sizes of the training set on the proposed method. To satisfy the conditions, the training data was divided into 5 folds (each fold include 20\% of the training set) randomly. In the next step, the training set was raised fold by fold each time and the impacts on MAcc improvement was observed. Table \ref{tab5} shows the influence of applying the different sizes of the training set to our system, with a fixed number of GMM components. This observation revealed better results in the first part of our experiments. The reported values in this table are based on the best results obtained from the different size of raw i-vectors and applying PCA and VAE to them. (In each case, results are reported using the parameters configuration for best results).\par
\begin{figure}[H]
\centering
\includegraphics[scale = 0.5]{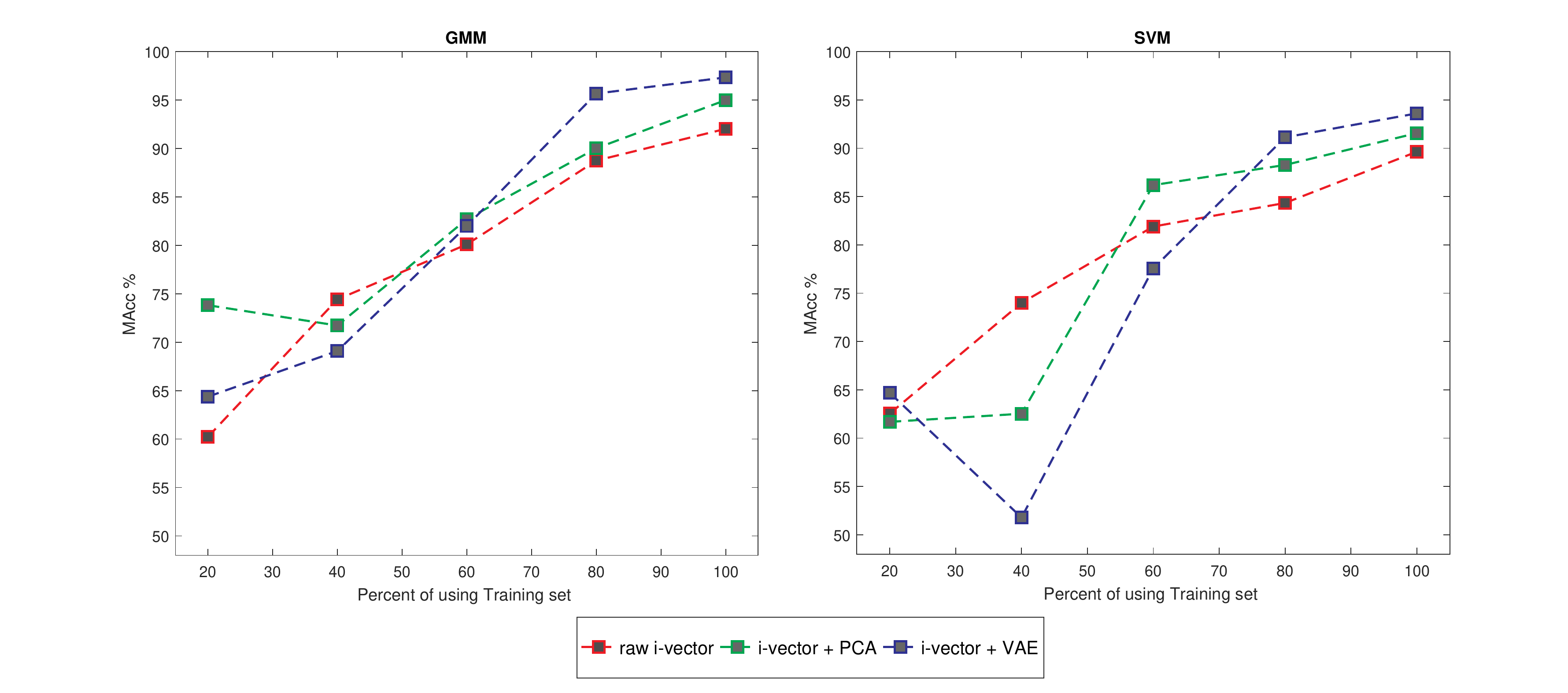}
\caption{DAT curve comparison for raw i-vector and its PCA and VAE for using different size of training set. In each case, results are reported using the best parameters configuration.}
\label{fig7}
\end{figure}

As summarized in Table.~\ref{tab5}, the classification performance improved by an increase in the amount of training data.
The results suggest that increasing the size of training data over 80\%, leads to less improvement, in comparison with the cases where the size of the training set is smaller.
According to Table.~\ref{tab5}, the performance of proposed system is similar to the baseline system when only 60\% of the training set is used for training the proposed system.
\begin{table}[htbp]
  \centering
  \scriptsize
  \caption{The Effect of Using Different Size of training set on the performance of the Proposed System with GMM and SVM-RBF}
    \begin{tabular}{cc|ccc|ccc}
    \hline
      &  & \multicolumn{3}{c|}{GMM Classfier} & \multicolumn{3}{c}{SVM Classifier} \\
    \hline
    \textbf{Size of training set} & \textbf{System}  & 
    \textbf{Se\%}  & \textbf{Sp\%} & \textbf{MAcc\%} &
    \textbf{Se\%}  & \textbf{Sp\%} & \textbf{MAcc\%}\\
    \hline
    \hline
    \multirow{3}{*}{20\%} & Raw i-vector & 
    86.00 & 34.44 & 60.22 & 87.25 & 37.84 & 62.54\\
     & i-vector + PCA & 
     95.33 & 52.32 & 73.82 & 74.66 & 48.72 & 61.69\\
     & i-vector + VAE & 
     40.00 & 88.74 & 64.37 & 58.93 & 71.00 & 64.96\\
     \hline
    \multirow{3}{*}{40\%} & Raw i-vector & 
    60.00 & 74.83 & 70.41 & 68.77 & 79.23 & 74.00\\
     & i-vector + PCA & 
     60.00 & 83.44 & 71.72 & 70.20 & 54.87 & 62.53\\
     & i-vector + VAE & 28.95 & 
     72.85 & 69.09 & 63.24 & 40.34 & 51.79\\
     \hline
    \multirow{3}{*}{60\%} & Raw i-vector & 
    65.33 & 94.7 & 80.10 & 68.57 & 95.24 & 81.90\\
     & i-vector + PCA & 
     82.00 & 83.44 & 82.70 & 85.55 & 86.83 & 86.19 \\
     & i-vector + VAE & 
     64.67 & 99.34 & 82.00 & 69.70 & 85.44 & 77.57\\
     \hline
    \multirow{3}{*}{80\%} & Raw i-vector & 
    89.33 & 87.42 & 88.75 & 85.62 & 83.06 & 84.34\\
     & i-vector + PCA & 
     88.00 & 92.05 & 90.02  & 87.34 & 89.20 & 88.27\\
     & i-vector + VAE & 
     93.33 & 98.01 & 95.67 & 90.12 & 92.15 & 91.13\\
     \hline
    \multirow{3}{*}{100\%} & Raw i-vector & 
    88.74 & 95.33 & 92.03 & 88.10 & 91.18 & 89.64\\
     & i-vector + PCA & 
     91.30 & 98.67 & 95.00 & 89.44 & 93.71 & 91.57\\
     & i-vector + VAE & 
     96.02 & 98.86 & 97.34 & 92.28 & 94.95 & 93.61\\
     \hline
     \end{tabular}%
  \label{tab5}
\end{table}
In addition, the performance of SVM is better than the GMM for lower amount of data. However, for more amount of training data, GMM system works better than SVM.
Fig.~\ref{fig7} depicts the impact of varying training set size on the MAcc values of the proposed system.\par
\textbf{Discussion:} As shown in these figures, the MAcc improves while the training set size is gradually increasing. Obviously, the number of the samples is crucial in improving the results, since it improves the generalization. Moreover, it helps the system to adapt to new samples. Nevertheless, the principal discussion is about the comparison of three different approaches we engaged to examine whether feature reduction is applicable or not. First, as it can be seen, using the larger dataset for raw i-vector demonstrates lower improvement compared with using PCA. Obviously, the dimension reduction in large scale and small dataset yields better performance than raw i-vector. The most critical point here is that VAE has the best performance. The major reason is that VAE, as one of Deep Neural Networks (DNNs), requires more data to generalize results and as data increases, the results for VAE improve over time. Hence, it yields the best results among all approaches. \par 
It is also worthy of note that GMM has a better improvement rate than SVM. The explanation behind this change is lies in the mechanism of each classifier. GMM uses data distribution to perform classification, while SVM is mostly dependent on feature space rather than data. Thus, increasing data results in more accurate data distribution and consequently, better impact on GMM performance. In addition, performance of SVM has a smooth and low improvement, because adding data in SVM results in better transformation through kernel function, not better feature space, necessarily.

\subsection{Best Results Analysis}
Table.~\ref{tab4} presents the results obtained by the baseline system and the best results obtained by our proposed systems in this study. Accordingly, the best MAcc is achieved by our proposed system which is 97.34\% when i-vector, VAE and GMM are employed. This result improves the accuracy of the baseline system by 15.84\%.\par
\begin{table}[htbp]
  \centering
  \scriptsize
  \caption{Best results for Our proposed approaches and baseline system.}
    \begin{tabular}{|c|c|c|c|c|}
    \hline
    \textbf{System} 
    & \textbf{Se\%}  & \textbf{Sp\%} & \textbf{MAcc\%} \\
    \hline
    baseline & 
    84.5 & 78.5 & 81.5 \\
    \hline
    \hline
    i-vector + GMM & 
    88.74 & 95.33 & 92.03 \\
    i-vector + PCA + GMM & 
    93.37 & 98.6 & 95.98 \\
    i-vector + VAE + GMM & 
    96.02 & 98.86 & 97.34 \\
    \hline
    i-vector + SVM & 
    88.10 & 91.18 & 89.64\\
    i-vector + PCA + SVM & 
    89.44 & 93.71 & 91.57\\
    i-vector + VAE + SVM & 
    92.28 & 94.95 & 93.61\\
    \hline
    \end{tabular}%
  \label{tab4}
\end{table}
\textbf{Discussion:} In the baseline system represented in ~\cite{ref56}, extracted features are mostly based on frequency and sub-band features; such as MFCC, Mel-Spectrogram, etc. These features are suitable for robust speech or sound detection. However, in other applications like heart sound classification, it is essential to extract an identical features for our purpose. This is due to specific characteristic of heart sound, that is unique for every individual. As a result, i-vector can be better features for heart sound identification. Hence, it can improve classification error and accuracy better than approaches based on robust feature extraction. In addition, GMM has superiority over SVM, since GMM gives a better description of samples in terms of feature space since this classifier obtains this goal with no change in feature space. On the other hand, SVM uses a kernel to map the current feature space to a better one and that can cause a problem, since it may be solved in the current feature space and changing the feature space can increase complexity.

\section{Conclusions}
This research study proposes a novel method for automatic heart sound classification based on i-vector MFCC features embedding. In this method, MFCC features are extracted from heart sounds, and then i-vectors are obtained based on these features. The achieved i-vectors represent the characteristics of the participants’ heart sound, given that a heart sound is unique for each individual. This method is based on fix-sized i-vector and therefore insensitive to the length of the input sounds. In addition, the i-vector of a heart sound is a more suitable feature to describe the characteristics of heart sound than other variable length features, since the whole sound is considered for i-vector producing. i-vectors are fed to PCA or VAE in order to produce an apt discrimination. Finally, these features are given to GMMs and SVM classifiers for final labeling. The experiments on a public dataset demonstrate the effectiveness of the proposed method. The combination of MFCC and i-vector is stable and can reflect the key point features to discriminate two types of the subject accurately. The proposed method also works well with limited amount of data. In conclusion, the proposed method outperforms the state-of-the-art approach.

\section{Acknowledgment}
We thank Mr. Mohammad Elmi and Mr. Majid Osati for comments that greatly improved the manuscript.


\newpage

\begin{wrapfigure}{l}{25mm} 
\includegraphics[width=1in,height=1.25in,clip,keepaspectratio]{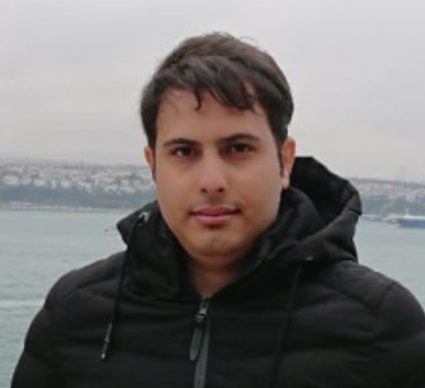}
\end{wrapfigure}\par
\textbf{Mohammad Adiban} was born in Babol, Iran, at 1991. He received his master in Artificial Intelligence (AI) from the Department of Computer Engineering, Sharif University of Technology (SUT) at 2017. He is currently a researcher at Speech Processing Laboratory (SPL) and his research interests include AI, Machine Learning, Audio Speech Processing, Natural Language Processing and Biomedical. \par
\vspace{1.5\baselineskip}
\begin{wrapfigure}{l}{25mm} 
\includegraphics[width=1in,height=1.25in,clip,keepaspectratio]{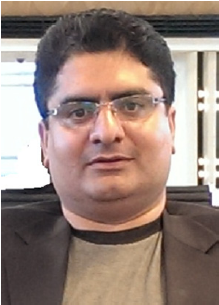}
\end{wrapfigure}\par
\textbf{Bagher BabaAli} is an Assistant Professor of Computer Science at University of Tehran since 2013. He received his B.S. degree in Computer Engineering from Shiraz University and his M.S. and PhD degrees both in Artificial Intelligence, from Sharif University of Technology, Iran, in 2003 and 2009, respectively. After his PhD degree, he has visited Johns Hopkins University in USA, Fondazione Bruno Kessler (FBK) in Italy, and CSIR in South Africa. He has served as a senior researcher of AGP, a speech and language technologies company established in Tehran, Iran, in 2004 where he played key role in the development of a large vocabulary speech recognition system for Persian language. His research interests span the areas of statistical machine learning and pattern recognition, sequential pattern labeling, and deep learning and he has authored and co-authored more than 70 scientific papers in these fields. \par
\vspace{1.5\baselineskip}
\begin{wrapfigure}{l}{25mm} 
\includegraphics[width=1in,height=1.25in,clip,keepaspectratio]{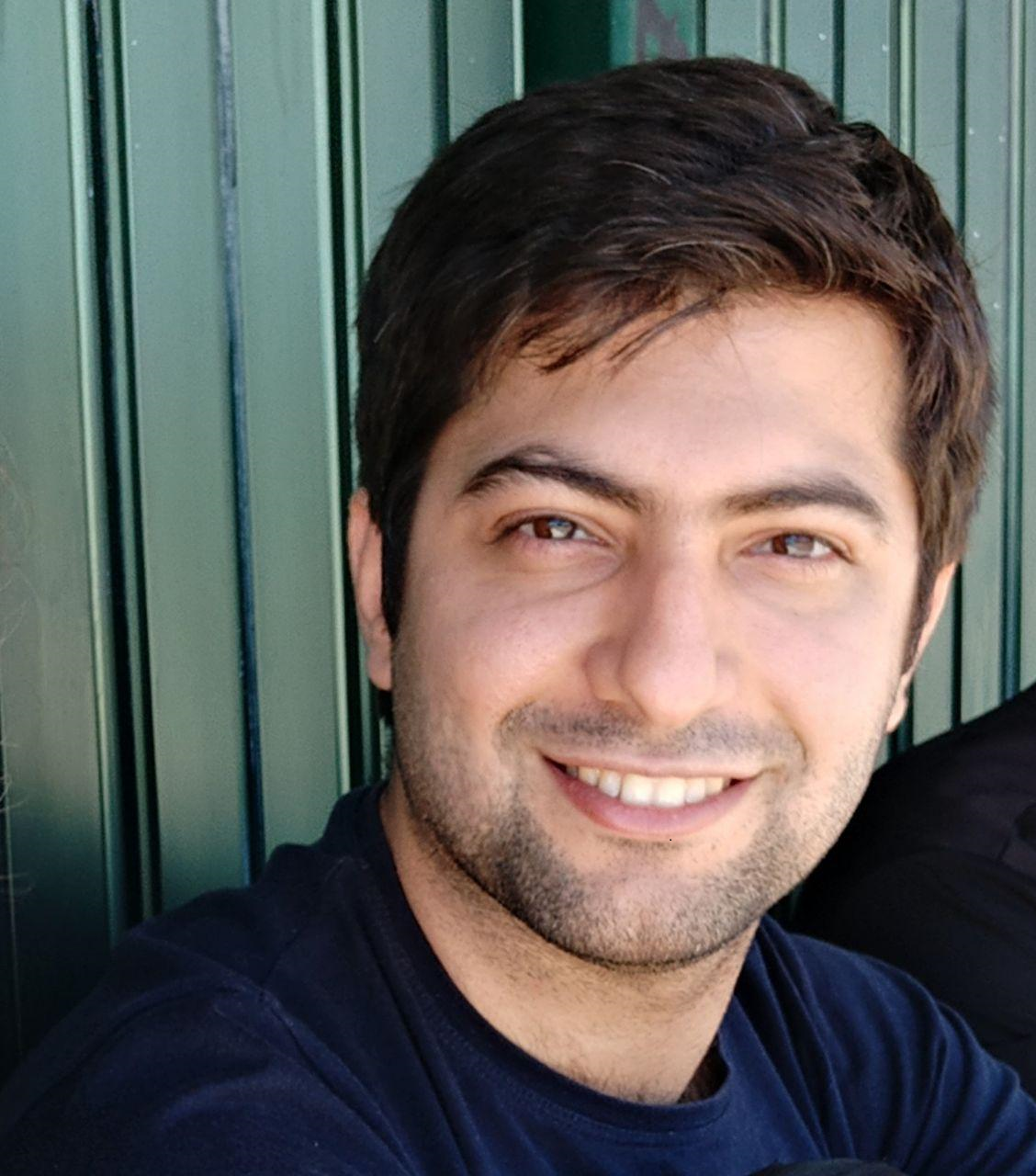}
\end{wrapfigure}\par
\textbf{Saeedreza Shehnepoor} was born in Yazd, Iran, at 1992. He received his BS from Sharif University of Technology, 2014 in Information Technology (IT). He finished his master in Artificial Intelligence and recieved MS degree from University of Tehran (UT) at 2016. He is currently a first year PhD student in University of Western Australia (UWA). His research field is concentrated on Machine Learning area and specifically issues related to spam contents in social media. \par

\end{document}